\documentclass[letterpaper, 10 pt, conference]{ieeeconf}  

\IEEEoverridecommandlockouts                              

\usepackage{graphics} 
\usepackage{epsfig} 
\usepackage{empheq}
\usepackage{times} 
\usepackage{amsmath} 
\usepackage{amssymb}  
\usepackage{bm}
\usepackage{color}
\usepackage{bbding} 
\usepackage{cite}
\usepackage{diagbox}
\usepackage[linesnumbered,ruled]{algorithm2e}
\usepackage{ulem} 
\usepackage{hyperref}
\hypersetup{
	colorlinks=true,
	linkcolor=blue,
	filecolor=blue,      
	urlcolor=blue,
	citecolor=cyan,
}
\usepackage{float}
\usepackage{booktabs}
\usepackage{multirow}
\usepackage{mathrsfs}
\usepackage[utf8]{inputenc}
\usepackage{subfigure}
\usepackage{graphicx}
\usepackage{pifont}
\usepackage{threeparttable}
\newcounter{RNum}
\renewcommand{\theRNum}{\arabic{RNum}}
\newcommand{\Remark}{\noindent\textbf{Remark}~\refstepcounter{RNum}\textbf{\theRNum}: }

\title{\LARGE \bf
	Predictive Visual Tracking: A New Benchmark and Baseline Approach
}

\author{Bowen Li$^{1, *}$, Yiming Li$^{2, *}$, Junjie Ye$^{1}$, Changhong Fu\textsuperscript{1, \ding{41}}, and Hang Zhao$^{3}$
	\thanks{* Equal contribution.}
	\thanks{\ding{41} Corresponding author.}
	\thanks{$^{1}$Bowen Li, Junjie Ye, and Changhong Fu are with School of Mechanical Engineering, Tongji University, Shanghai, China {\tt\small changhongfu@tongji.edu.cn}}%
	\thanks{$^{2}$Yiming Li is with Tandon School of Engineering, New York University, USA {\tt\small yimingli@nyu.edu}}%
	\thanks{$^{3}$Hang Zhao is with Institute of Interdisciplinary Information Sciences, Tsinghua University, Beijing, China {\tt\small hangzhao@tsinghua.edu.cn}}%
}

\begin{document}
	\maketitle
	\thispagestyle{empty}
	\pagestyle{empty}
	
	\begin{abstract}
		As a crucial robotic perception capability, visual tracking has been intensively studied recently. In the real-world scenarios, the onboard processing time of the image streams inevitably leads to a discrepancy between the tracking results and the real-world states. However, existing visual tracking benchmarks commonly run the trackers offline and ignore such latency in the evaluation. In this work, we aim to deal with a more realistic problem of latency-aware tracking. The state-of-the-art trackers are evaluated in the aerial scenarios with new metrics jointly assessing the tracking accuracy and efficiency. Moreover, a new predictive visual tracking baseline is developed to compensate for the latency stemming from the onboard computation. Our latency-aware benchmark can provide a more realistic evaluation of the trackers for the robotic applications. Besides, exhaustive experiments have proven the effectiveness of the proposed predictive visual tracking baseline approach. Our code is on ~\url{https://github.com/vision4robotics/LAE-PVT-master}.
	\end{abstract}

	\section{Introduction}
	Visual tracking\footnote{We target single object tracking in this work.}, one of the fundamental perception tasks for mobile robots, has introduced various real-world applications~\cite{Bonatti2019IROS2,Fu2014ICRA,Li2020AutoTrackTH}. In the past decade, visual tracking benchmarks~\cite{Wu2015ObjectTB,Mueller2016ECCV,Fan2019LaSOTAH,Huang2019GOT10kAL} have generally focused on the offline evaluation, \textit{i.e.}, they query the state of the object on each image frame with a ground-truth annotation. In the real world, however, when the algorithm finishes processing a captured frame, the surrounding environment has changed in varying degrees, \textit{i.e.}, \textit{the output of the tracker is always outdated}. 
	
	Especially for Unmanned Aerial Vehicle (UAV) tracking applications, such latency can be more detrimental due to: 1) the scarce onboard computation resource ends up in longer processing time, \textit{i.e.}, more remarkable latency, 2) the fast motion scenarios on UAV cause drastic state changes of the target objects in a very short time. Consequently, the delayed tracking results can jeopardize the robot perception and planning modules. In visual tracking community, many attempts are made to improve the tracking accuracy and robustness on the existing offline benchmarks~\cite{Li2018HighPV, Li2019SiamRPNEO, Danelljan2019ATOMAT, Bhat2019LearningDM}. Although the performance has been largely improved, a latency-aware benchmark/approach jointly considering the accuracy as well as latency is more desirable for the robotics community. 
	
	\begin{figure}[t]
		\centering
		\includegraphics[width=0.95\columnwidth]{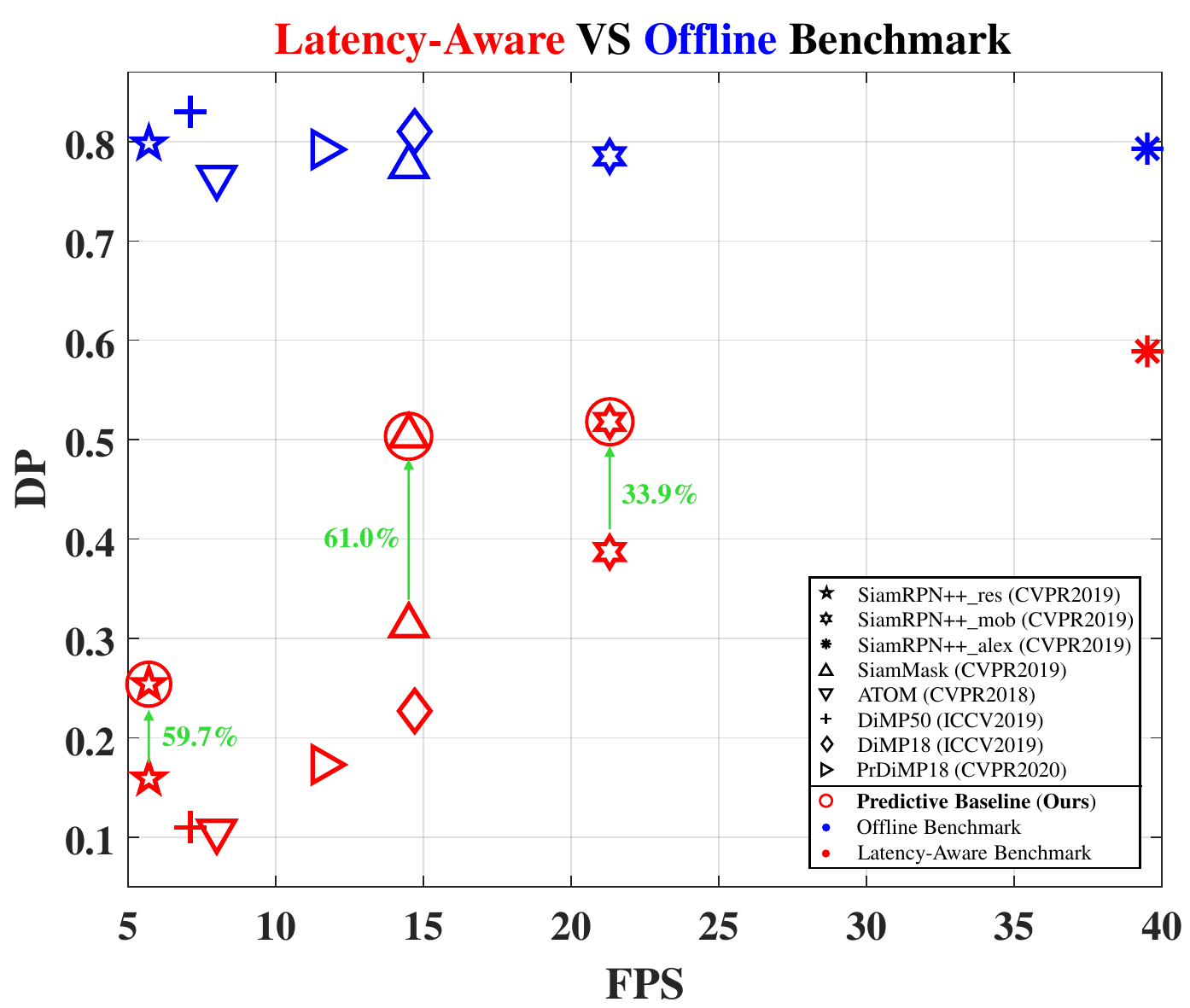}
		\vspace{-0.2cm}
		\caption{Performance of the state-of-the-art trackers with offline latency-free and online latency-aware benchmarks on DTB70 \cite{li2017AAAI}. The distance precision (DP) is employed for evaluation. The same shape indicates the same tracker, \textit{e.g.}, star for SiamRPN++~\cite{Li2019SiamRPNEO} with ResNet50 as backbone. \textbf{\textcolor[rgb]{0,0,1}{Blue}} denotes the results on the offline benchmark. \textbf{\textcolor{red}{Red}} means the results on the latency-aware benchmark. Our predictive tracking baseline is marked out by red circles, where the original performance is improved by a considerable margin, denoted by \textbf{\textcolor[rgb]{0.2,0.86,0.2}{green}} arrows and percentages.}
		\label{fig:1}
		\vspace{-0.4cm}
	\end{figure}
	
	In this work, different from the offline evaluation querying the state of the object on each image, we propose a \textit{latency-aware evaluation} (LAE) approach by querying the state of the target object at each timestamp. In our settings, the onboard computation would introduce extra time, therefore, \textit{the trackers have to output the future states of the object, to compensate for the latency}. Essentially, visual tracking has become a joint perception and forecasting task, whereas the offline tracking is solely a perception mission. We name such a novel task as \textit{predictive visual tracking} (PVT). Compared to the offline tracking, PVT is more realistic and can yield the states of the object in real time, yet suffers from increased uncertainties and is more challenging. 
	
	In the context of predictive visual tracking, we firstly benchmark several state-of-the-art trackers~\cite{wang2019Mask, Li2019SiamRPNEO, Danelljan2019ATOMAT, Bhat2019LearningDM, Martin2020Pr} against the latency-aware criteria on a high-end UAV processor. As illustrated in Fig.~\ref{fig:1}, compared to the offline latency-free benchmark (in blue color), the tracking performance is greatly lowered in the latency-aware settings (in red color), especially for the trackers with low speed. Moreover, we propose a novel and effective plug-in prediction module (in red circles). Our contributions are three-fold:
	
	\begin{itemize}
		\item A novel task of joint tracking and forecasting, i.e., predictive visual tracking (PVT), is proposed to bridge the gap between computer vision and real-world robotics.
		\item A new latency-aware evaluation (LAE) benchmark is developed to comprehensively evaluate the trackers in terms of both accuracy and efficiency. 
		\item Exhaustive experiments have been extended on an onboard device to point out the urgent latency issue and validate the proposed PVT baseline.
	\end{itemize}
	
	To the best of our knowledge, we are the first to establish a realistic latency-aware benchmark in visual tracking. The results in Fig.~\ref{fig:1} indicate the significance for the robotics and vision community to focus on PVT. We release our code and toolkit to facilitate further investigations for the community. Important notations used for deduction are shown in TABLE~\ref{tab:1} for clear reference.
	
	\section{Related Works}
	
	\subsection{Visual Tracking Approaches}
	Visual tracking aims to localize the object of interest in the image streams given solely the information of the target on the first frame. Current methods generally consider the tracking as a confidence regression problem: given the filter/template and the current frame, cross correlation is implemented and the location with the highest confidence score will be the output. Discriminative correlation filter (DCF)-based~\cite{Henriques2015TPAMI,Galoogahi2017ICCV,Danelljan2017CVPR,Mueller2017CVPR,Huang2019LearningAR, Li2020AutoTrackTH} and Siamese network-based trackers~\cite{Li2018HighPV, Li2019SiamRPNEO, Danelljan2019ATOMAT, Bhat2019LearningDM} are two main paradigms in visual tracking community. DCF-based trackers are highly efficient thanks to the fast Fourier transform (FFT) and are frequently used in robotics applications~\cite{Li2020AugmentedMF,Li2020KeyfilterAwareRU,Li2020TrainingSetDF,He2020TowardsRV,Fu2020DR2TrackTR}. Yet there is still an obvious gap between their performance and that of deep learning methods. Siamese trackers have excellent performance but is much slower especially on the mobile robots with constrained calculation resources. In this work, we benchmark the advanced deep trackers in the latency-aware scenarios and try to improve the performance via a predictive baseline.
	
	\subsection{Visual Tracking Benchmarks}
	Since OTB was proposed in 2013~\cite{Wu2015ObjectTB}, many visual object tracking datasets have been developed following the evaluation metrics in OTB, such as LaSOT~\cite{Fan2019LaSOTAH}, TrackingNet~\cite{Mller2018TrackingNetAL}, GOT-10K~\cite{Huang2019GOT10kAL}, \textit{etc}., in order to improve the diversity of the tracking scenarios. Moreover, multi-modality tracking~\cite{Liu2020PTBTIRAT, Lukezic2019CDTBAC}, first-person view tracking~\cite{Dunnhofer2020IsFP}, day-and-night tracking~\cite{li2021All}, aerial tracking~\cite{Mueller2016ECCV, li2017AAAI} were also proposed to further boost the real-world applications of visual tracking. In addition to OTB protocol, VOT~\cite{Kristan2017TheVO,Kristan2018TheSV,Kristan2019TheSV}, an annual visual object tracking challenge, has employed a re-initialization scheme when encountering tracking failure. In a word, existing tracking benchmarks generally address the tracking accuracy and robustness in various tracking scenes, while neglecting the evaluation of the real-world latency introduced by the computation. In this work, we reformulate the evaluation protocol in visual tracking by jointly assessing the tracking accuracy and latency. We employ the aerial tracking benchmarks~\cite{Mueller2016ECCV,li2017AAAI,Du2018ECCV} captured by the real-world UAV for realistic evaluations.
	
	\subsection{Latency in Perception}
	In the computer vision community, the problem of latency has been extensively studied. \cite{Wang2020RealTimeCT,howard2017mobilenets,sandler2018mobilenetv2} proposed lightweight architectures to ease the computation. \cite{Li2020AutoTrackTH,Huang2019LearningAR} proposed efficient algorithms to decrease the running time. Researchers generally claim that the algorithm is real-time when the running speed is higher than the frame rates. However, even for the algorithms faster than the frame rates, the output is still not real-time since the state of the world has already changed when a specific frame is processed. \cite{Li2020TowardsSP} instantiated a benchmark of streaming object detection to highlight the importance of the latency in the real-time perception. Yet there is still no latency-aware benchmark and baseline approach in the area of visual object tracking, where latency issue is actually more critical.
	
	\begin{table}[!t]
		\centering
		\setlength{\tabcolsep}{0.9mm}
		\fontsize{6}{11.2}\selectfont
		\caption{List of the important notations in this work.}
		\begin{tabular}{ccc}
			\toprule[1.5pt]
			\textbf{Symbol}	& \textbf{Meaning}  & \textbf{Dimension}  \\
			\midrule
			$i$ & World frame number & $\mathbb{R}$\\
			$T$ & Sequence length & $\mathbb{R}$\\
			$k$ & Serial number of the processed frame  & $\mathbb{R}$\\
			$j_k$ & Input frame id for the tracker (World frame id of the processed frame) & $\mathbb{R}$\\
			$K$ & Total number of the processed frames& $\mathbb{R}$\\
			$t_{\mathrm{g}}^{i}$ & World timestamp & $\mathbb{R}$\\
			$t_{\mathrm{r}}^{j_k}$ & Tracker timestamp & $\mathbb{R}$\\
			$\mathcal{T}^{j_k}$ & Processing time of frame $j_k$ & $\mathbb{R}$\\
			$\psi(i) (=j_h)$ & Input frame id to be paired with frame $i$ & $\mathbb{R}$\\
			$\Delta t_k$ & Time interval between two frames $j_{k+1}$ and $j_k$ & $\mathbb{R}$\\
			$\Delta t_i$ & Time interval between two frames $i$ and $\psi{i}$ & $\mathbb{R}$\\
			$\Delta t_h$ & Time interval between two frames $j_{h+1}$ and $j_h$ & $\mathbb{R}$\\
			$\mathbf{b}_{\mathrm{g}}^{i}$ & Ground-truth bounding box in frame $i$ & $\mathbb{R}^{1\times4}$\\
			$\mathbf{b}_{\mathrm{r}}^{j_k}$ & Raw output bounding box in frame $j_k$ & $\mathbb{R}^{1\times4}$\\
			$\hat{\mathbf{b}}_{\mathrm{r}}^{i}$ & Final output bounding box to be evaluated & $\mathbb{R}^{1\times4}$\\
			$\mathbf{s}_{j_k}$ & Updated object state in frame $j_k$ & $\mathbb{R}^{8\times1}$\\
			$\hat{\mathbf{s}}_{j_k}$ & Predicted object state in frame $j_k$ & $\mathbb{R}^{8\times1}$\\
			$\mathbf{z}_{j_{k+1}}$ & Measured (Tracked) object box in frame $j_{k+1}$ & $\mathbb{R}^{4\times1}$\\
			$\mathbf{P}_{j_k}$ & Updated covariance matrix in frame $j_k$ & $\mathbb{R}^{8\times8}$\\
			$\hat{\mathbf{P}}_{j_k}$ & Predicted covariance matrix in frame $j_k$ & $\mathbb{R}^{8\times8}$\\
			$\mathbf{F}(\Delta t)$ & Transition matrix for time interval $\Delta t$ & $\mathbb{R}^{8\times8}$\\
			$\mathbf{Q}(\Delta t)$ & Noise matrix for time interval $\Delta t$ & $\mathbb{R}^{8\times8}$\\
			$\mathbf{K}_{j_k}$ & Kalman gain for frame $j_k$ & $\mathbb{R}^{4\times4}$\\
			$\mathbf{H}$ & Slicing matrix & $\mathbb{R}^{4\times8}$\\
			$\mathbf{I}_N$ & Identity matrix & $\mathbb{R}^{N\times N}$\\
			\bottomrule[1.5pt]
		\end{tabular}%
		\label{tab:1}%
		\vspace{-0.4cm}
	\end{table}%
	\section{Latency-Aware Benchmark}
	
	\subsection{Ground-truth Reformulation}
	To query the object's state at each timestamp, we firstly define a world frame rate $\kappa$ set as 30 frames/s (FPS). So the ground-truths $\mathbf{G}=[\mathbf{g}_0,\mathbf{g}_1,\dots,\mathbf{g}_{T-1}]\in\mathbb{R}^{6\times T} $ are:
	
	\begin{equation}
	\mathbf{g}_i=[i,\mathbf{b}_\mathrm{g}^i,t_\mathrm{g}^i]^{\mathrm{T}}~,
	\end{equation}
	\noindent where $i$ and $T$ denote frame number and sequence length respectively, $\mathbf{b}_\mathrm{g}^i=[x_i,y_i,w_i,h_i]$ indicates the ground-truth bounding box in the $i$-th frame, and $t_\mathrm{g}^i=\frac{i}{\kappa}$ is the world timestamp.

	\subsection{Tracking Schedule}\label{Sec:3B}
	
	In the real-world tracking scenarios, a tracker faster than the world frame rate is able to process every captured frame. However, slow trackers with large latency can simply process the latest frame, discarding the frames captured during the calculation. The output results $\mathbf{R}=[\mathbf{r}_0,\mathbf{r}_1,\dots,\mathbf{r}_{K-1}]\in\mathbb{R}^{6\times K} $ of the tracker can be formulated as:
	\begin{equation}
	\mathbf{r}_k=[j_k,\mathbf{b}_\mathrm{r}^{j_k},t_\mathrm{r}^{j_k}]^{\mathrm{T}}~,
	\end{equation}
	where  $k$ denotes the $k$-th processed frames with a length of $K$ and $j_k$ denotes the world frame number. $\mathbf{b}_\mathrm{r}^{j_k}$ represents the output bounding box on frame ${j_k}$. $t_\mathrm{r}^{j_k}$ means the tracker timestamp, \textit{i.e.,} the time after the tracker finishes processing frame ${j_k}$. Specifically, define $\mathcal{T}^{j_k}$ as the time consumed by the tracker to process frame ${j_k}$, $t_\mathrm{r}^{j_k}$ can be calculated as:
	\begin{equation}\label{eqn:3}
	t_\mathrm{r}^{j_k}=\sum_{x=0}^{k}\mathcal{T}^{j_x}~.
	\end{equation}
	
	In this case, the latency-aware output $\mathbf{R}$ depends on ${j_k}$, \textit{i.e.}, which frame to be processed. This work considers ${j_k}$ as the latest available frame number as:
	\begin{equation}\label{eqn:4}
	\mathcal{I}_f, \mathcal{I}_0, ,\mathcal{I}_1, , \mathcal{I}_2, , \mathcal{I}_3 , \mathcal{I}_4
	\end{equation}
	\Remark Due to the latency, the processed frame id $j_k$ could be nonconsecutive, \textit{i.e.}, $\mathbf{j}=[j_0,j_1,\dots,j_{K-1}]\subseteq[0,1,\dots,T-1]$. Consequently, the motion between frame $j_k$ and $j_{k+1}$ is much stronger, making latency-aware tracking more challenging $\mathcal{H}_{\phi(11)}$.
	
	\subsection{Evaluation Process}\label{sec:3C}
	
	\begin{figure}[t]
		\centering
		\includegraphics[width=0.95\columnwidth]{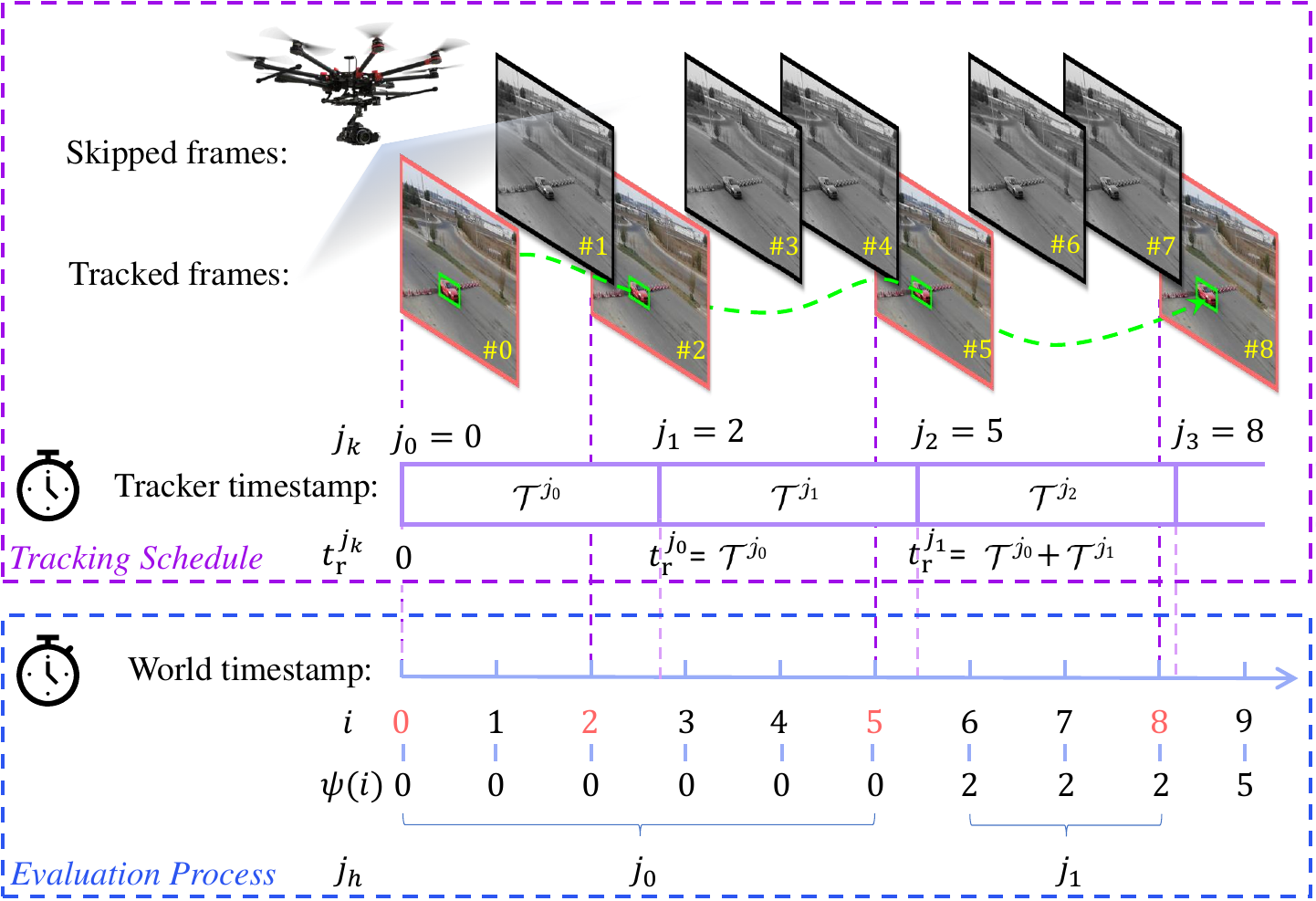}
		\caption{Example of the proposed latency-aware benchmark. If the tracker is slower than the world frame rate, some frames would be skipped. Also, there is a mismatch between the evaluated frame $i$ and the latest tracked frame $\psi(i)$. The tracking schedule and evaluation metric are exhibited in purple and blue dotted lines, respectively.}
		\label{fig:2}
		\vspace{-0.4cm}
	\end{figure}
	Different from offline benchmark pairing the ground-truth in the $i$-th frame with the output in the $i$-th frame, the proposed LAE benchmark pairs ground-truth $\mathbf{b}^i_\mathrm{g}$ at time $t_\mathrm{g}^i$ with the latest output result $\mathbf{b}_\mathrm{r}^{\psi(i)}$ where $\psi(i)$ is:
	\begin{equation}\label{eqn:5}
	\psi(i) = \left\{
	\begin{array}{crl}
	0 &, & {t_\mathrm{g}^i < t_\mathrm{r}^{j_0}}\\
	\mathop{\arg\max}_{j_k}\ t_\mathrm{r}^{j_k}\le t_\mathrm{g}^i &, & {\mathrm{others}}
	\end{array}\right.~.
	\end{equation}
	\Remark In practice, there is no ``in time" tracking results. Since $\mathcal{T}^{j_k}>0$, the output time for frame $i$ will definitely satisfy $t_\mathrm{r}^{i}>t_\mathrm{g}^i$. This makes the latest input frame id $\psi(i)<i\ (i>0)$. Hence, the output will be at least one frame behind the ground-truth. Figure~\ref{fig:2} illustrated the overall picture of the processing and evaluation in our LAE benchmark.

	
	\section{Predictive Visual Tracking}
\begin{figure}[t]
		\centering
		\includegraphics[width=1\columnwidth]{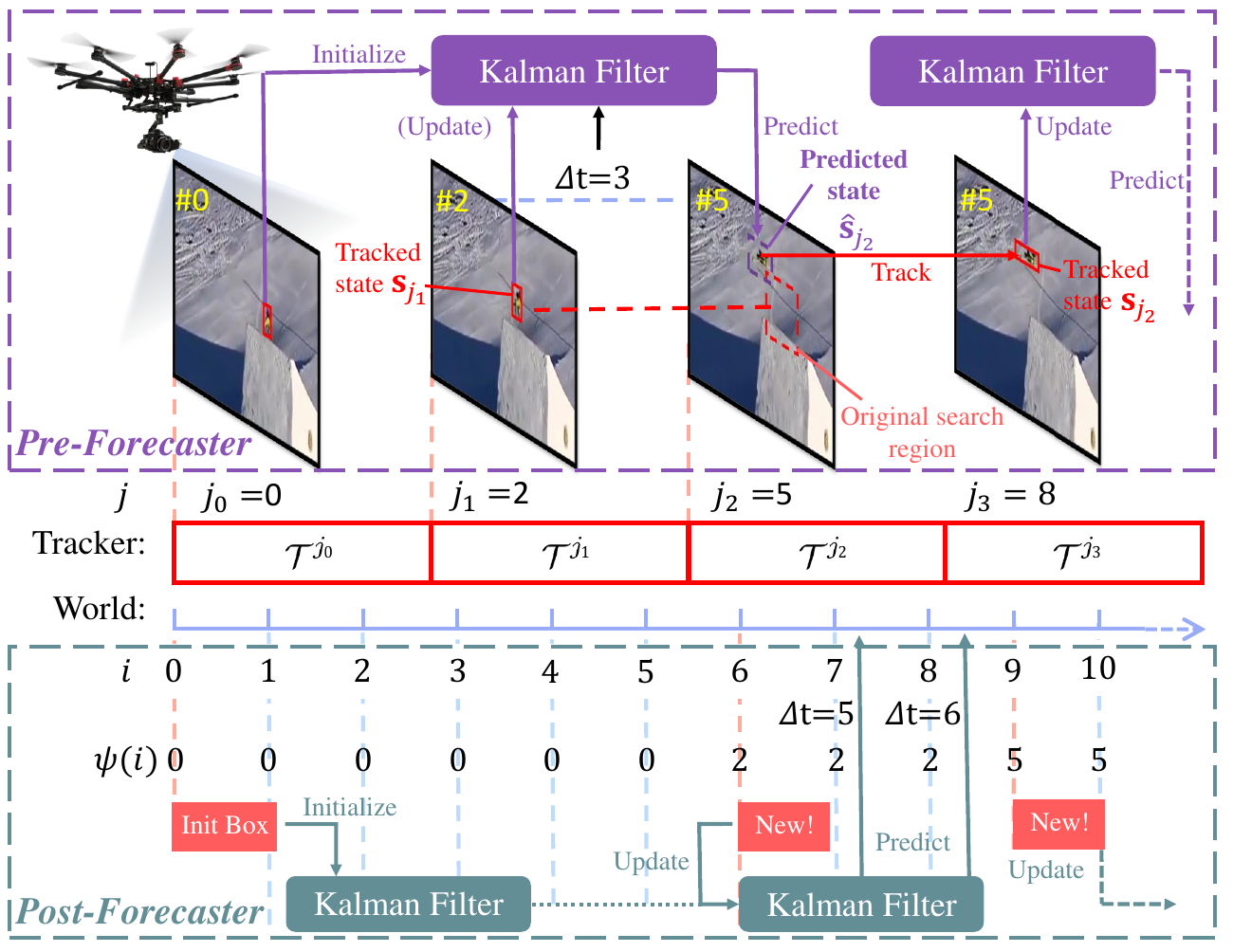}
		\vspace{-0.3cm}
		\caption{Visualization of the proposed predictive visual tracking framework. The red elements denote tracking procedure, the purple elements indicate pre-forecasting process, and the blue elements represent the post-forecaster. }
		\label{fig:main}
		\vspace{-0.3cm}
	\end{figure}

	This section will introduce our PVT baseline framework which contains two forecasters based on the first-order Kalman Filter (KF) \cite{kalman1960}. As is shown in Fig.~\ref{fig:main}, the pre-forecaster can make up for the skipped-frames and predict the search region. The post-forecaster aims at remedying the mismatch between the evaluated and processed frames. For clearer understanding, Alg.~\ref{alg:PVT} illustrates how the proposed PVT works during online tracking.
	
	\subsection{Pre-Forecasting of Search Region}
	As is pointed out in Sec.~\ref{Sec:3B}, the number of discarded frames could be very large because of the low tracking speed. In visual tracking framework, the search region for input frame $j_{k}$ is usually centered around the result of the previous processed frame $j_{k-1}$. If two successive input frames $j_{k-1}$ and $j_{k}$ stride too wide, the target object would has a large displacement and could be out of the search region, leading to tracking failure. Therefore, the  pre-forecaster is proposed to predict the search region before the correlation operation.
	
	Consider the optimal world state at frame $j_k$ as a vector $\mathbf{s}_{j_k}=[x_{j_k},y_{j_k},w_{j_k},h_{j_k},\dot{x}_{j_k},\dot{y}_{j_k},\dot{w}_{j_k},\dot{h}_{j_k}]^{\mathrm{T}}$, where $[x_{j_k},y_{j_k},w_{j_k},h_{j_k}]$ is the coordinate of the left top point, width as well height of the bounding box. The corresponding covariance matrix is $\mathbf{P}_{j_k}\in\mathbb{R}^{8\times8}$. Then, the predicted state in frame ${j_{k}}$ and the updated covariance matrix are:
	\begin{equation}\label{eqn:6}
	\begin{aligned}
	&\hat{\mathbf{s}}_{j_{k}}&=&\ \mathbf{F}(\Delta t_k)\mathbf{s}_{j_{k-1}}~,\\
	&\hat{\mathbf{P}}_{j_{k}}&=&\ \mathbf{F}(\Delta t_k)\mathbf{P}_{j_{k-1}}\mathbf{F}(\Delta t_k)^{\mathrm{T}}+\mathbf{Q}(\Delta t_k)~,
	\end{aligned}
	\end{equation}
	\noindent where $\mathbf{F}(\Delta t)$ and $\mathbf{Q}(\Delta t)$ respectively denotes the transition matrix and the process noise covariance matrix in time interval $\Delta t$. During tracking, the latency accumulation makes the time interval $\Delta t_k=j_{k}-j_{k-1}$ between two successive processed frames variable. Thus, the Kalman Filter should be time-varying. The transition and noise matrix are:
	\begin{equation}\label{eqn:7}
	\begin{aligned}
	\mathbf{F}(\Delta t)&=\begin{bmatrix}
	\mathbf{I}_4  &  \Delta t\mathbf{I}_4  \\
	\mathbf{0}  &  \mathbf{I}_4 
	\end{bmatrix}~,\\
	\mathbf{Q}(\Delta t)&=(\Delta t)^2\mathbf{I}_8~,
	\end{aligned}
	\end{equation}
	\noindent where $\mathbf{I}_4\in\mathbb{R}^{4\times4}$ denotes identity matrix. Intuitively, the process noise is larger in longer time intervals. When frame $j_{k}$ arrives and the measured object state $\mathbf{z}_{j_{k}}=[x_{j_{k}},y_{j_{k}},w_{j_{k}},h_{j_{k}}]^{\mathrm{T}}$ is available, the Kalman Filter follows the calculation below for update:
	\begin{equation}\label{eqn:8}
	\begin{aligned}
	&\mathbf{K}_{j_{k}}&=&\ \hat{\mathbf{P}}_{j_{k}}\mathbf{H}^{\mathrm{T}}(\mathbf{H}\hat{\mathbf{P}}_{j_{k}}\mathbf{H}^{\mathrm{T}}+\mathbf{R})^{-1}~,\\
	&\mathbf{s}_{j_{k}}&=&\ \hat{\mathbf{s}}_{j_{k}}+\mathbf{K}_{j_{k}}(\mathbf{z}_{j_{k}}-\mathbf{H}\hat{\mathbf{s}}_{j_{k}})~,\\
	&\mathbf{P}_{j_{k}}&=&\ (\mathbf{I}-\mathbf{K}_{j_{k}}\mathbf{H})\hat{\mathbf{P}}_{j_{k}}~,
	\end{aligned}
	\end{equation}
	
	\noindent where $\mathbf{H}=[\mathbf{I}_4,\mathbf{0}]\in\mathbb{R}^{4\times8}$ is the slicing operation and $\mathbf{R}$, equaling to $10\times\mathbf{I}_4$, remains constant. Hereafter, when a new frame $j_{k+1}$ arrives, the forecaster firstly adopts Eq.~(\ref{eqn:6}) to get the predicted object state $\hat{\mathbf{s}}_{j_{k+1}}$ to correct the search region. After the tracker finishes processing frame $j_{k+1}$ and obtains the measured object state $\mathbf{z}_{j_{k+1}}$, the forecaster is updated using Eq.~(\ref{eqn:8}), then waiting for the arrival of frame $j_{k+2}$.
	
	\begin{algorithm}[!t]
		\caption{Predictive Visual Tracking}
		\label{alg:PVT}
		\KwIn {A video sequence of $T$ frames\\
			\hspace{0.95cm} Init bounding box $\mathbf{b}_{\mathrm{g}}^0$ of the tracked object \\}
		\KwOut {Paired estimated object bounding box $\hat{\mathbf{b}}^{i}_{\mathrm{r}}$ in all upcoming frames}
		Initialize world timestamp $t^i_\mathrm{g}$\\
		Initialize the result box $\hat{\mathbf{b}}_{\mathrm{r}}^0=\mathbf{b}_{\mathrm{g}}^0$\\
		\For{frame number $i=0$ to end}{ 
			\eIf{$i=0$}{
				Tracker and forecasters initialization with $\mathbf{b}_{\mathrm{g}}^0$\\
				Update $t^0_\mathrm{r}$ using Eq.~(\ref{eqn:3})\\
			}
			{
				Find $j_k$ using Eq.~(\ref{eqn:4})\\
				\eIf{$j_{k}$ is new}
				{
					Calculate $\Delta t_k=j_k-j_{k-1}$\\
					Predict $\hat{\mathbf{s}}_{j_{k}}$ with $\Delta t_k$ and $\mathbf{s}_{j_{k-1}}$ by Eq.~(\ref{eqn:6})\\
					Correct search region with $\hat{\mathbf{s}}_{j_{k}}$\\
					Tracking in the corrected search region\\
					Update pre-forecaster with $\mathbf{s}_{j_k}$ by Eq.~(\ref{eqn:8})\\
					$k++$\\
				}
				{
					Wait for the next new frame\\
				}
			}
			Find $\mathbf{b}_{\mathrm{r}}^{\psi(i)}$ using Eq.~(\ref{eqn:5}) (suppose $\psi(i)=j_h$)\\
			\If{$\psi(i)>0$}{
				Calculate $\Delta t_{\psi(i)}=i-\psi(i)$\\
				\If{$\psi(i)$ is new}{
					Calculate $\Delta t_h=j_h-j_{h-1}$\\
					Update post-forecaster with $\Delta t_h$ by Eq.~(\ref{eqn:6}) and Eq.~(\ref{eqn:8})\\}
				Predict $\hat{\mathbf{s}}_i$ using Eq.~(\ref{eqn:9})\\
				Obtain paired output box $\hat{\mathbf{b}}^{i}_{\mathrm{r}}=\mathbf{H}\hat{\mathbf{s}}_i$\\
				Update tracker timestamp $t^{j_k}_\mathrm{r}$ using Eq.~(\ref{eqn:3})\\
			}
		}
	\end{algorithm} 
	
	\subsection{Post-Forecasting of Object States}
	As is illustrated in Sec.~\ref{sec:3C}, the latest input frame id $\psi(i)$ is smaller than the evaluated frame $i$. To compensate for the discrepancy between $\psi(i)$ and $i$, the post-forecaster basically follows the same work-flow of the pre-forecaster. Yet the update and prediction of KF in the post-forecaster are implemented asynchronously. Specifically, the prediction is carried out at each evaluation frame $i$, while the update is implemented only when $\psi(i)$ renovates. For a clear explanation, let's suppose $\psi(i)=j_h$ below according to Eq.~(\ref{eqn:5}). For the prediction, the time interval $\Delta t_{\psi(i)}=i-\psi(i)$ is used to generate $\mathbf{F}(\Delta t_{\psi(i)})$ using Eq.~(\ref{eqn:7}). The predicted state $\hat{\mathbf{s}}_{i}$ and final output bounding box to be evaluated of frame $i$ is obtained using:
	\begin{equation}\label{eqn:9}
	\hat{\mathbf{s}}_{i}=\mathbf{F}(\Delta t_{\psi(i)})\mathbf{s}_{j_h}~,
	\hat{\mathbf{b}}^{i}_{\mathrm{r}}=\mathbf{H}\hat{\mathbf{s}}_{i}~.
	\end{equation}
	
	When a new output $\mathbf{z}_{j_{h+1}}$ which corresponds to the input frame id $j_{h+1}$ becomes the latest processed frame, the time interval between 2 different raw output id $j_h$ and $j_{h+1}$ is $\Delta t_{h}=j_{h+1}-j_{h}$. With generated $\mathbf{F}(\Delta t_{h})$ and $\mathbf{Q}(\Delta t_h)$ using Eq.~(\ref{eqn:7}), the post-forecaster adopts the same strategy to update as the pre-forecaster, \textit{i.e.}, Eq.~(\ref{eqn:6}) and Eq.~(\ref{eqn:8}), with different input id $j_h$ rather than $j_k$.
	
	\Remark The speed of KF-based forecasters is about 500 FPS on a cheap CPU, and our prediction module is a generic framework which can be efficiently embedded into any tracker to improve the performance.
	
	\section{Experiments}
	
	\begin{figure*}[!t]
		\centering
		\includegraphics[width=2\columnwidth]{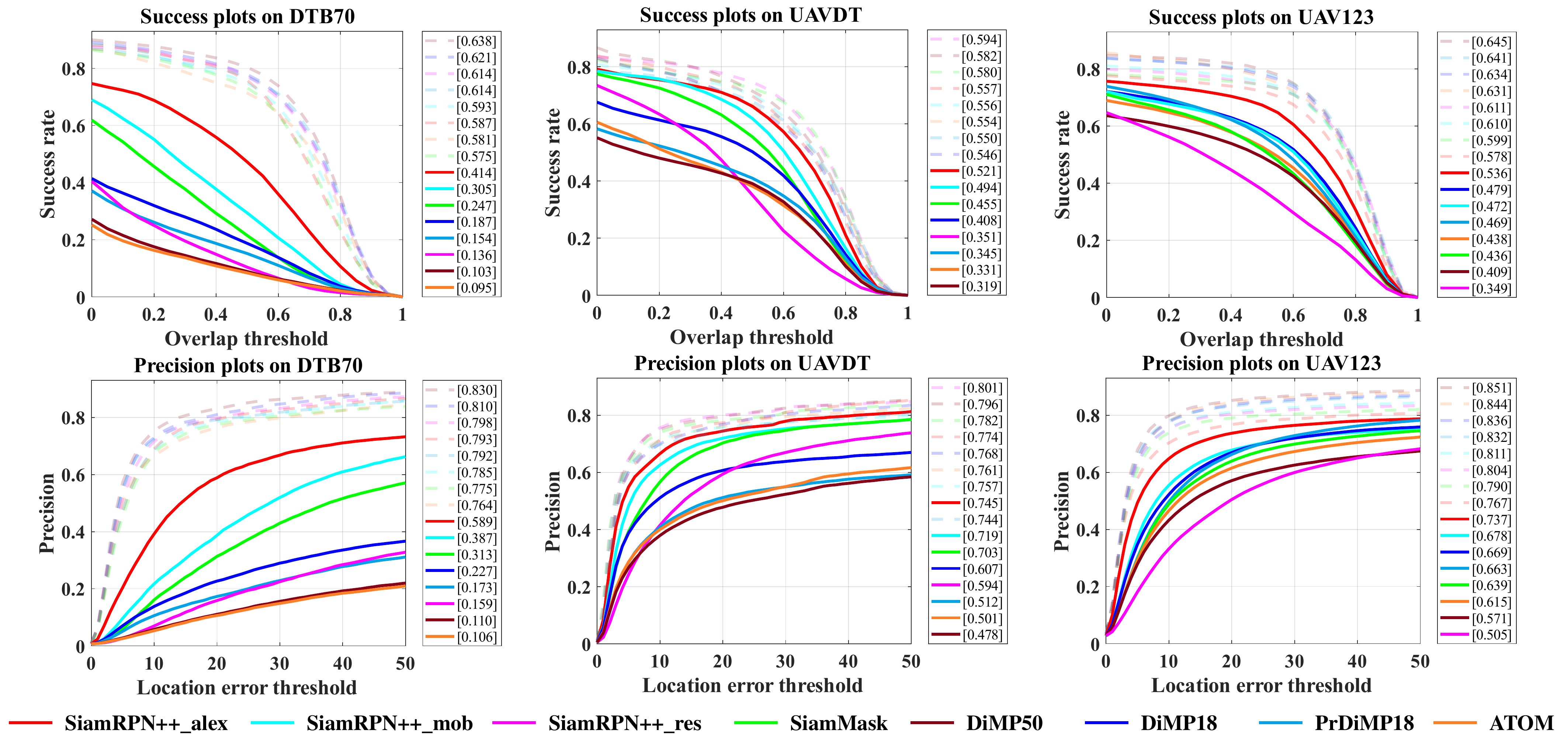}
		\vspace{-0.2cm}
		\caption{Performance of state-of-the-art trackers on the proposed LAE benchmark. The curves in solid colors report the performance of the 8 benchmarked trackers on LAE, whereas the dotted curves overlaid in semi-transparent same colors outline the performance obtained by the same trackers on the traditional offline benchmark. In brackets, we report the distance precision (DP) and area under curve (AUC) on LAE (in black) and on offline benchmark (in gray). Clearly, many offline promising trackers fail to maintain their robustness and accuracy in LAE benchmark. }
		\label{fig:overall}
	\end{figure*}
	
	\begin{table*}[!t]
		\centering
		\setlength{\tabcolsep}{1mm}
		\fontsize{6}{7}\selectfont
		\caption{Effect of the proposed PVT baseline on traditional SOTA trackers. The improvement in percentage and running speed are also reported for reference. \Checkmark indecates whether the proposed 2 forecasters are enabled. Under most cases, the forecasters are able to boost the original performance of the SOTA trackers in new LAE benchmark.}
		\begin{tabular}{cccccccccccccccccccccc}
			\toprule
			Datasets & \multicolumn{7}{c}{DTB70}                             & \multicolumn{7}{c}{UAVDT}                             & \multicolumn{7}{c}{UAV123} \\
			\cmidrule{1-22}    Metric & \multicolumn{3}{c}{AUC} & \multicolumn{3}{c}{DP} & FPS   & \multicolumn{3}{c}{AUC} & \multicolumn{3}{c}{DP} & FPS   & \multicolumn{3}{c}{AUC} & \multicolumn{3}{c}{DP} & FPS \\
			\cmidrule(lr){2-4}\cmidrule(lr){5-7}\cmidrule(lr){9-11}\cmidrule(lr){12-14}\cmidrule(lr){16-18}\cmidrule(lr){19-21}   Forecasters & \XSolidBrush & \Checkmark & $\Delta$(\%) & \XSolidBrush & \Checkmark & $\Delta$(\%) &       & \XSolidBrush & \Checkmark & $\Delta$(\%) & \XSolidBrush & \Checkmark & $\Delta$(\%) &       & \XSolidBrush & \Checkmark & $\Delta$(\%) & \XSolidBrush & \Checkmark & $\Delta$(\%) &  \\
		    SiamRPN++\_res & 0.136 & 0.201 & \textbf{+47.8} & 0.159 & 0.254 & \textbf{+59.7} & 6.0   & 0.351 & 0.467 & \textbf{+33.0} & 0.594 & 0.687 & \textbf{+15.7} & 6.0   & 0.349 & 0.434 & \textbf{+24.3} & 0.505 & 0.605 & \textbf{+19.8} & 5.6 \\
			SiamRPN++\_mob & 0.305 & 0.377 & \textbf{+23.6} & 0.387 & 0.518 & \textbf{+33.9} & 21.2  & 0.494 & 0.533 & \textbf{+7.9} & 0.719 & 0.74  & \textbf{+2.9} & 22.5  & 0.472 & 0.522 & \textbf{+10.6} & 0.678 & 0.722 & +6.5  & 22.0 \\
			SiamMask & 0.247 & 0.362 & \textbf{+46.6} & 0.313 & 0.504 & \textbf{+61.0} & 14.8  & 0.455 & 0.539 & \textbf{+18.5} & 0.703 & 0.751 & +6.8  & 15.3  & 0.436 & 0.514 & \textbf{+17.9} & 0.639 & 0.701 & \textbf{+9.7} & 14.6 \\
			\bottomrule
		\end{tabular}%
		
		\label{tab:overall}%
	\end{table*}%

	\subsection{Implementation Details}
	\subsubsection{Evaluation Platform}
	All the experimental evaluation in this work adopts NVIDIA Jetson AGX Xavier, a typical processing device onboard UAV.
	\subsubsection{Evaluation Metric}
	
	This work adopts generally-used metrics in visual tracking for evaluation. The first is distance precision (DP) based on the center location error between the output box and the ground-truth box. The second is area under curve (AUC) represented by the intersection ratio of the predicted box and the ground-truth box. 
	
	\Remark Different from the offline benchmarks on which the tracking speed is separated from robustness, LAE actually considers the trade-off between efficiency and robustness of a tracker. A slow tracker may be extremely precise in offline benchmarks, while the large latency will make its performance drop significantly in LAE scenarios. 
	\subsubsection{Evaluation Datasets}
	We use three UAV tracking datasets for a comprehensive evaluation, \textit{i.e.}, DTB70 \cite{li2017AAAI}, UAVDT \cite{Du2018ECCV}, and UAV123 \cite{Mueller2016ECCV} (all recorded at 30 FPS). Note that with the same datases, different benchmarks, \textit{i.e.}, offline latency-free benchmark and the proposed LAE benchmark are adopted while evaluation.
	
	\subsection{Results on the LAE Benchmark}
	We employ eight SOTA trackers, \textit{i.e.}, SiamRPN++\_res, SiamRPN++\_mob, SiamRPN++\_alex\footnote{\_res, \_mob, and \_alex respectively denote the 3 different backbones adopted, \textit{i.e.}, Resnet50, Mobilenetv2, and Alexnet.} \cite{Li2019SiamRPNEO}, SiamMask \cite{wang2019Mask}, ATOM \cite{Danelljan2019ATOMAT}, DiMP18, DiMP50 \cite{Bhat2019LearningDM}, and PrDiMP18 \cite{Martin2020Pr}, to assess their performances in real-world tracking applications. Figure~\ref{fig:overall} exhibits the AUC and DP on traditional offline benchmark (curves in semi-transparent dash lines and values in gray) and the newly proposed LAE benchmark (curves in solid colors and values in black). In summary, all the SOTA trackers undergo a severe performance decline in LAE scenarios compared to the offline evaluation. For the fast trackers like SiamRPN++\_alex \cite{Li2019SiamRPNEO}, the performance drop caused by the latency is slightly smaller than the slow trackers like SiamRPN++\_res \cite{Li2019SiamRPNEO}. For instance, in DTB70 \cite{li2017AAAI}, the DP of SiamRPN++\_alex \cite{Li2019SiamRPNEO} is lowered by about 22.9\% in LAE, while for SiamRPN++\_res \cite{Li2019SiamRPNEO}, the drop can be up to \textbf{76.7}\%. Such results have proven the importance of the tracking efficiency in the real-world latency-aware tracking. For DTB70 \cite{li2017AAAI} with strong UAV motion, the trackers are faced with larger performance drop. Besides, the initialization time can also lead to non-negligible latency for the near-real-time trackers. Our LAE jointly assessing tracking accuracy and efficiency can provide a more realistic evaluation for the practical tracking applications.
	
	
	
	\subsection{Predictive Visual Tracking Framework}

	\subsubsection{Overall Evaluation}
	The predictive framework is implemented on four SOTA trackers, \textit{i.e.}, SiamRPN++\_res, SiamRPN++\_mob, and SiamMask \cite{wang2019Mask}.  TABLE~\ref{tab:overall} presents the overall results on three datasets, where the DP, AUC, improvements in percentage, and running speed of the original trackers and predictive trackers are exhibited. In most cases, the predictive tracker can achieve far better results than its original framework with a gain of more than \textbf{20\%}. Besides, the slower the tracker runs, the larger improvements are achieved by the PVT framework. Take DTB70 \cite{li2017AAAI} 
	as an example, the PVT framework brings SiamRPN++\_res \cite{Li2019SiamRPNEO} up by more than \textbf{40\%} in both DP and AUC. In terms of a faster tracker SiamRPN++\_mob \cite{Li2019SiamRPNEO}, the improvement is less remarkable (around \textbf{30\%}). Nevertheless, our PVT framework cannot work well on SiamRPN++\_alex \cite{Li2019SiamRPNEO} with a real-time frame rate. The main reason is that the original tracker is fast enough to process each frame, so the original search region can be appropriately employed while the use of a pre-forecaster can be misleading.
	
	\Remark Coupled with PVT, the non-real-time trackers can now achieve a comparable results than the real-time trackers lik eSiamRPN++\_alex.
	
	\subsubsection{Attribute-Based Evaluation}
	The performance under ten aerial tracking attributes (scale variation (SV), aspect ratio variation (ARV), occlusion (OCC), deformation (DEF), fast camera motion (FCM), in-plane rotation (IPR), out-of-plane rotation (OPR), out-of-view (OV), similar objects around (SOA), and motion blur (MB)) are reported in TABLE~\ref{tab:att}. The proposed PVT has improved the robustness of the trackers notably, ranking the first in all attributes, \textit{e.g.}, in FCM, our tracker SimRPN++\_mob\_f outperforms its baseline by \textbf{57\%} and in the DEF 
	scenario, our SiamMask\_f nearly doubles its baseline. The attributes represent the typical challenges of the real-world UAV tracking compared with the general static object tracking, \textit{i.e.}, larger appearance variation and fast motion. In the LAE, the above difficulties are largely raised since some frames could be skipped. The attribute-based results have proven the promising effect of PVT in UAV tracking, alleviating the influence caused by the latency.
	
	\begin{table}[!t]
		\centering
		\setlength{\tabcolsep}{0.82mm}
		\fontsize{5.5}{7}\selectfont
		\caption{DP results in DTB70 \cite{li2017AAAI} with LAE benchmark by UAV special attributes. \textcolor[rgb]{ .753,  0,  0}{\textbf{Red}} denotes first place. }
		\begin{tabular}{ccccccccccc}
			\toprule[1.5pt]
			\diagbox{Trackers}{Att.} & ARV   & DEF   & FCM   & IPR   & MB    & OCC   & OPR   & OV    & SV    & SOA \\
			\midrule
   			ATOM  & 0.116 & 0.070 & 0.079 & 0.101 & 0.060 & 0.140 & 0.071 & 0.076 & 0.114 & 0.091 \\
			DiMP18 & 0.214 & 0.158 & 0.197 & 0.22  & 0.153 & 0.256 & 0.135 & 0.236 & 0.287 & 0.210 \\
			DiMP50 & 0.127 & 0.083 & 0.082 & 0.111 & 0.055 & 0.155 & 0.077 & 0.236 & 0.124 & 0.112 \\
			PrDiMP18 & 0.154 & 0.150 & 0.136 & 0.142 & 0.100 & 0.197 & 0.136 & 0.071 & 0.210 & 0.146 \\
			SiamMask & 0.247 & 0.246 & 0.241 & 0.236 & 0.167 & 0.361 & 0.128 & 0.222 & 0.287 & 0.34 \\
			SiamRPN++\_mob & 0.34  & 0.374 & 0.363 & 0.349 & 0.246 & 0.408 & 0.213 & 0.413 & 0.366 & 0.333 \\
			SiamRPN++\_res & 0.101 & 0.083 & 0.106 & 0.124 & 0.051 & 0.223 & 0.083 & 0.062 & 0.123 & 0.217 \\
			\midrule
			SiamMask\_f (ours) & \textcolor[rgb]{ .753,  0,  0}{\textbf{0.517}} & \textcolor[rgb]{ .753,  0,  0}{\textbf{0.535}} & 0.424 & 0.444 & 0.347 & 0.513 & \textcolor[rgb]{ .753,  0,  0}{\textbf{0.399}} & 0.355 & \textcolor[rgb]{ .753,  0,  0}{\textbf{0.55}} & \textcolor[rgb]{ .753,  0,  0}{\textbf{0.445}} \\
			SiamRPN++\_mob\_f (ours) & 0.427 & 0.429 & \textcolor[rgb]{ .753,  0,  0}{\textbf{0.572}} & \textcolor[rgb]{ .753,  0,  0}{\textbf{0.482}} & \textcolor[rgb]{ .753,  0,  0}{\textbf{0.483}} & \textcolor[rgb]{ .753,  0,  0}{\textbf{0.409}} & 0.219 & \textcolor[rgb]{ .753,  0,  0}{\textbf{0.498}} & 0.501 & 0.436 \\
			SiamRPN++\_res\_f (ours) & 0.227 & 0.241 & 0.181 & 0.206 & 0.087 & 0.286 & 0.133 & 0.18  & 0.203 & 0.295 \\
			\bottomrule[1.5pt]
		\end{tabular}%
		\label{tab:att}%
	\end{table}%
	
	\subsubsection{Ablation Study}
	To break down the performance gains from each part of PVT, ablation study is conducted on SiamRPN++\_mob baseline~\cite{Li2019SiamRPNEO}. As is shown in TABLE~\ref{tab:abla}, each forecaster is embedded into the tracker individually. The results have illustrated that both forecasters can yield a performance gain on the original tracking framework (for the pre-forecaster, the improvement is about \textbf{11\%} and for the post-forecaster, it is around \textbf{20\%}). The final version equipped with both forecasters achieved the best result, up to \textbf{23.6\%} in AUC and \textbf{33.9\%} in DP.
	
	\begin{table}[!t]
		\centering
		\caption{Ablation study of the 2 forecasters on SiamRPN++\_mob \cite{Li2019SiamRPNEO}. \Checkmark indicates whether the corresponding forecaster is enabled. Clearly, each forecaster can make an improvement and the final results with both forecasters are the best.}
		\begin{tabular}{ccccc}
			\toprule
			Tracker & Pre-Forecaster & Post-Forecaster & AUC   & DP \\
			\midrule
			\multirow{6}[1]{*}{Baseline} &       &       & 0.305 & 0.387 \\
			\cmidrule{2-5}
			& \Checkmark     &       & 0.318 & 0.430 \\
			\cmidrule{2-5}
			&       & \Checkmark     & 0.349 & 0.482 \\
			\cmidrule{2-5}
			& \Checkmark     & \Checkmark     & \textbf{0.377} & \textbf{0.518}\\
			\bottomrule
		\end{tabular}%
		\label{tab:abla}%
	\end{table}%
	\begin{figure}[!t]
		\centering
		\vspace{-0.1cm}
		\includegraphics[width=1\columnwidth]{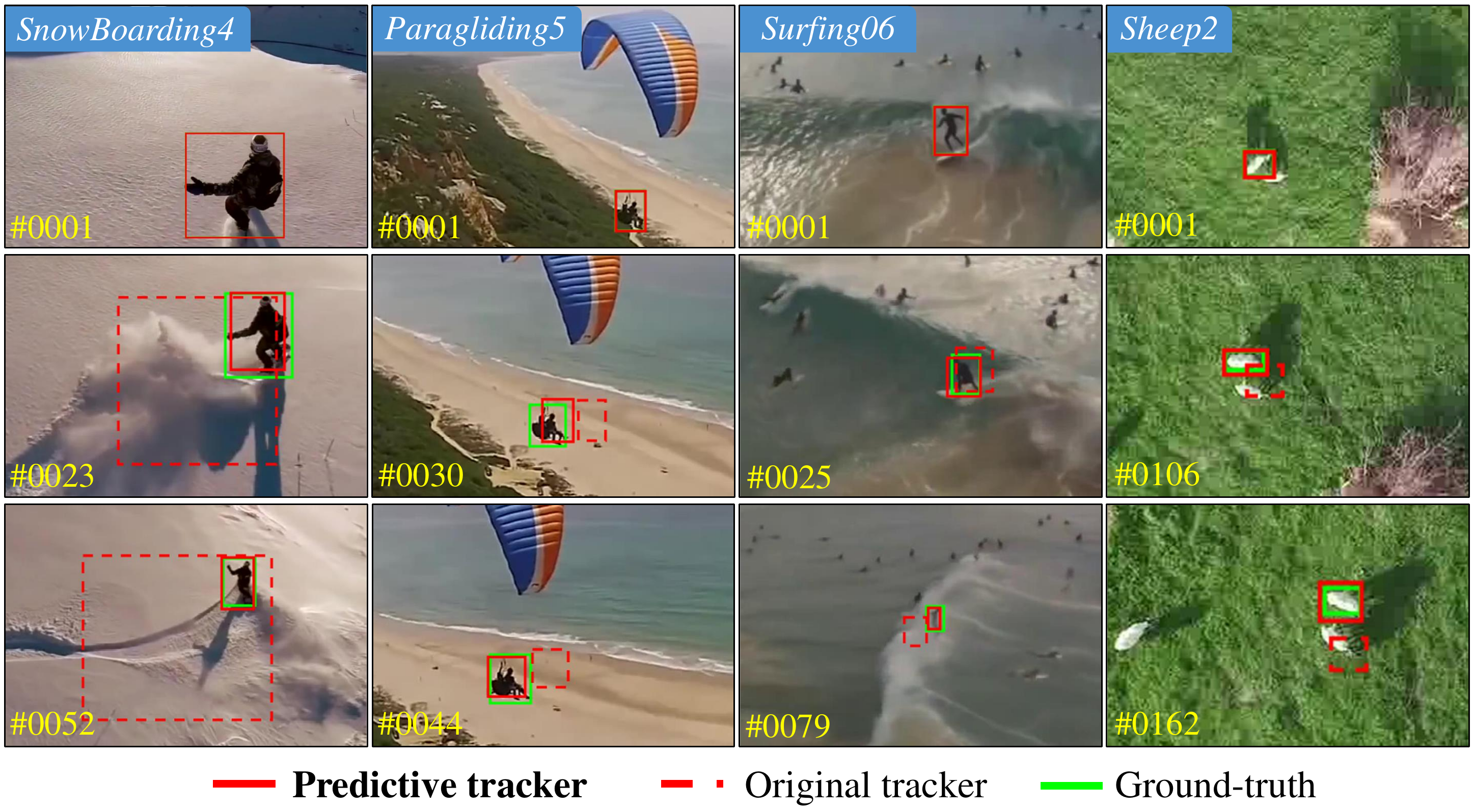}
		\vspace{-0.3cm}
		\caption{Visualization of representative latency-aware tracking scenes. The \textcolor{red}{red} boxes denotes bare tracker (dashed line) and proposed predictive tracker (solid line). The \textcolor[rgb]{0.2,0.86,0.2}{green} boxes indicates ground truth. More latency-aware tracking sequences can be found at \url{https://youtu.be/n8i8bREIFeM}. }
		\label{fig:vis}
		\vspace{-0.4cm}
	\end{figure}
	\subsubsection{Qualitative Evaluation}
	Figure~\ref{fig:vis} showed several tracking examples from DTB70~\cite{li2017AAAI}, adopting LAE benchmark. In the real-world latency-aware tracking, the bare trackers (in dashed red boxes) easily fall invalid, while the proposed PVT framework (in solid red boxes) has improved the robustness by a large margin.
	
	\section{Conclusions}
	We propose a brand-new task of predictive visual tracking along with a pioneering latency-aware evaluation (LAE) method for visual tracking community. The proposed LAE has formulated both accuracy and efficiency into a holistic evaluation process, to provide a realistic evaluation of the trackers for the robotics applications. Moreover, a generic and effective predictive tracking framework is proposed to compensate for the latency. The comprehensive experiments not only demonstrate the promising effect of our baseline approach, but also  point out the demand to investigate latency-aware PVT for the robotics and computer vision community. We strongly believe our work can promote the applications of visual object tracking in the real world.
	


	\section*{Acknowledgment}
	This work is supported by the National Natural Science
	Foundation of China (No. 61806148) and the Natural Science
	Foundation of Shanghai (No. 20ZR1460100).

	\bibliographystyle{IEEEtran}  
	\normalem
	\bibliography{IEEEabrv,ref}
	
\end{document}